\documentclass{uai2023} 

\usepackage[american]{babel}

\usepackage{subfigure}

\usepackage{geometry}
\usepackage{amsmath}
\newtheorem{definition}{Definition}[section]
\usepackage{mathrsfs}
\usepackage{bm}
\usepackage[cmintegrals]{newtxmath}
\usepackage{float}
\usepackage{subfigure}

\usepackage{color}
\usepackage{framed}
\definecolor{shadecolor}{rgb}{0.92,0.92,0.92}

\usepackage{booktabs}
\usepackage{multirow}
\usepackage{caption} 

\usepackage{enumerate}
\usepackage{algorithm}
\usepackage{algorithmic}

\newtheorem{hyp}{Hypothesis}

\title{GDBN: a Graph Neural Network Approach to Dynamic Bayesian Network}

\author[1]{\href{mailto:<sunyang.46135@bytedance.com>?Subject=Your UAI 2023 paper}{Yang Sun}{}}
\author[1]{Yifan Xie}
\affil[1]{
    Bytedance, P.R.China
}
  
\begin{document}
\maketitle

 \begin{abstract}
Identifying causal relations among multi-variate time series is one of the most important elements towards understanding the complex mechanisms underlying the dynamic system. It provides critical tools for forecasting, simulations and interventions in science and business analytics.
In this paper, we proposed a graph neural network approach with score-based method aiming at learning a sparse DAG that captures the causal dependencies in a discretized time temporal graph. We demonstrate methods with graph neural network significantly outperformed other state-of-the-art methods
with dynamic bayesian networking inference. In addition, from the experiments, the structural causal model can be more accurate than a linear SCM discovered by the methods
such as Notears.
\end{abstract}

\section{Introduction}

Multi-variate time series data has been carefully studied for many years. Sensors have been recording large amount of collected time series data including traffics, electricity consumption and so on. Interestingly, these time series variables are subtly internally interlinked. These links with causal dependencies carry critical information for decision makers.

\begin{figure}[!h]  
    \centering
		\subfigure[]{
			\centering
			\includegraphics[height=1.0in]{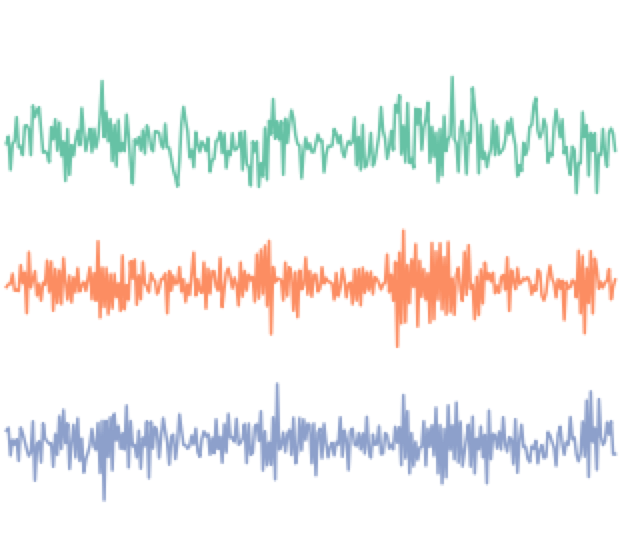}
		}
		\quad
		\subfigure[]{
			\centering
			\includegraphics[height=1.0in]{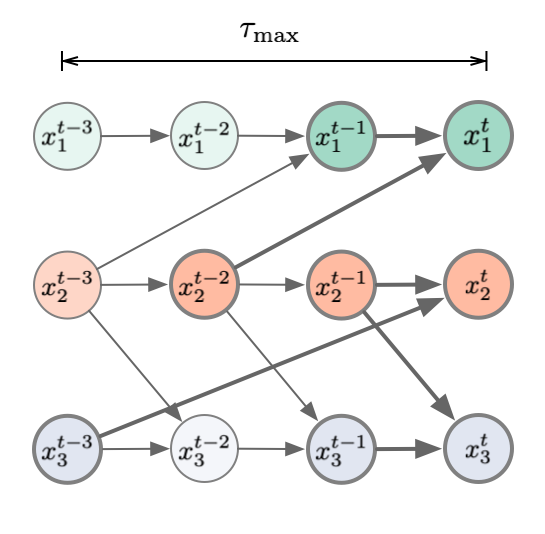}
		}
		\quad
		\subfigure[]{
			\centering
			\includegraphics[height=1.0in]{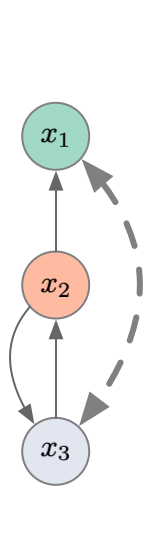}
		}
	\caption{An example of time series (a) with the corresponding full time graph (b) and summary graph (c) under our assumptions. The bold nodes and arrays in (b) constitute a causal temporal graph (Definition \ref{def:ctg}), which highlights the connections between variables at a time slice $t$ with their past. The maximum time lag $\tau_{\max}$ is 3. $x_2$ influence $x_3$ with lag=1, which is denoted as $a_{32}^1$. There exist spurious associations represented by gray dashed arrows in (c), which could lead to errors of misidentification of causal relations.} 
  \label{FTG}
\end{figure}

Causality has been formalized as a theory for causal and counterfactual inference \cite{pearl_2009} from a structural model. Once knowledge about the causality among time series variables is not a priori, we have to discover these relational structure. Moreover, the dynamics described from a structural equation model can articulate the causal dependencies via a directed acyclic graph (DAG). Causal models with explicitly solved DAG enables computation of posterior distribution with treatment\cite{Varian7310}, \cite{shalit2017estimating} and provides us with actionable insights. A careful treatment of causal relation will also avoid misleading judgement from misidentifying spurious correlations as illustrated from Figure \ref{FTG}.

Recently, research done on time series data mainly focuses on the summary graph \cite{lowe2020amortized} \cite{rodas2021causal} for multi-variate time series. We are more interested with latent structure that governs the multi-variate time series data. The other closedly related topic is graph neural network. Deep neural network has been known as universal function approximator for data distribution and a powerful tool for representation learning. Graph neural network is known as the specific class of deep neural network that is suitable for graph-structured data. With a clear definition of interactions (with directions) among time series data with lags, we recast our problem of Bayesian Dynamics Networks as a link prediction problem via Graph Neural Network (GNN) \cite{yu2019dag}. The adjacency matrix of the DAG is provided as an input to the graph neural network which will be the objective of our inference problem. One of the most closely related work is \cite{pamfil2020dynotears} based on the linear version of Notears. However the generation process of real world time series data is highly complex and it is difficult to describe them as the usual linear relations.

\hfill

The main contributions of ours can be summarized in the following.
\begin{itemize}

\item By understanding the intrinsic property of the time series data, we define the causal temporal graph and its associated adjacency matrix. We introduce the fundamental elements used in our causal discovery framework.

\item  We synthesize a wide range of linear dataset and non-linear dataset from a generation process of VAR(p) with stationary property for causal discovery. 

\item  We further designed a graph-based neural network with a neural message passing scheme that caters for the intrinsic property of multi-variate time series. The graph structure is optimized based on
a variational autoencoder structure which generalizes better on understanding the distribution of time series data. We formulate the optimization objective with variational lower bound and sparsity constraint to solve for causal temporal graph and the SEM.

\item We carried out experiments over our synthesized data of different settings and datasets published on \cite{Lawrence2021DataGP}. Experiments show that our approach significantly outperformed methods from linear method as Notears \cite{zheng2018dags} and conditional independence framework \cite{runge2019detecting}.

\end{itemize}

\section{Related Work}


The causal inference problem and causal feature selection for time series data has been an issue under different contexts, including econometric, biology, physics and etc. Granger Causality \cite{Granger1969} based on prediction method has been proposed as a corner stone methodology for time series causal analysis. It used a variation of the VAR(p) model and the lagging causal effect is modeled with different matrices with index p. Despite its advantages, it suffers from the often lack of the causal sufficiency condition in real scenario. Lower sampling rate on the time dimension could also lead to unidentifiability of the Granger Causality, although some exceptional cases are studied \cite{pmlr-v37-gongb15}. An alternative is the PC algorithm that tests the conditional independence along with the autocorrelation \cite{Runge} \cite{runge2019detecting}. Despite its capacity of accurate causal discovery, this type of methods usually scale with number of variables and $\tau_{max}$ in polynomial time. For example \cite{runge2019detecting} shows the time complexity is $N^3 \tau^2_{max}$ where $N$ is the number of variables while the complexity of our inference method is of $N^2 \tau_{max}$. \cite{pmlr-v139-mastakouri21a} gives the necessary and sufficient condition to select time series random variable with conditional dependencies based on certain assumptions for the time series data. \cite{besserve2021causeeffect} introduces spectral method for identifying causal effect with robustness to down-sampling. 

Score-based methods formulate the causal graph inference problem with an objective function which regularizes the adjacency matrix to satisfy the constraint with respect to being a DAG \cite{huang-score}. NO TEARS\cite{zheng2018dags} recasts the combinatoric graph search problem as a continuous optimization problem, while DYNOTEARS\cite{pamfil2020dynotears} provides a vector-autoregressive extension to multivariate time series data. More recently, \cite{rethinking} derived a first principle relation between Graph Neural Network and SCM. Our approach is based on \cite{yu2019dag} where the authors treat the problem of DAG inference as a link prediction problem.

Neural ODE \cite{NEURIPS2018_69386f6b} has started a new chapter for understanding the dynamics of multi-variate time series data in continuous time domain
 \cite{chen2019neural} had revealed a fundamental situation where ordinary differential equation can be discovered and \cite{bellot2022neural} approximates a vector fields parametrized by a neural network for modelling the continuous dynamical system. \cite{pmlr-v151-de-brouwer22a} further develops the techniques for conditional treatment effect estimation.

\section{Problem Formulation}
\subsection{Assumptions}\label{s1} 
We assume Markov property and faithfulness holds, and that there are no instantaneous effects or hidden common causes. Then identifiabilty is guaranteed \cite{Malinsky2018CausalSL, peters2017elements}. Our causal discovery work is based on the assumption used in \cite{pmlr-v139-mastakouri21a}. Besides, the causal relations are  invariant under a joint time shift of all variables.

\noindent We represent the observations of the multi-variate time series $X$ as $\big( X_1^t, ..., X_d^t \big)$ with discrete timestamps.

\noindent Following Mastakouri et al.\cite{pmlr-v139-mastakouri21a}, we further assume, in the process of data generation,

\begin{hyp}[H\ref{hyp:first}] \label{hyp:first}
Lag is one for univariate time series: $x_i^{t-\tau}$ influence $x_i^t$ if and only if $\tau=1,\forall i\in\mathbb{N^*}$.
\end{hyp}

\begin{hyp}[H\ref{hyp:second}] \label{hyp:second}
Single-lag:There is only one lag for each pair of time series $(x_i, x_j),\forall i,j\in\mathbb{N^*}$.
\end{hyp}

\noindent where $\tau$ denotes the time lag between two temporal nodes.

\subsection{Definitions}

 \begin{definition}\label{def:ctg}
 causal temporal graph: \textnormal{Causal temporal Graph $G = \{V, E\}$ is an equivalence class up to time translational invariance. It is composed of vertices $V = \{    {X^t, X^{t-1},\cdots, X^{t-\tau_{max}}}  \}$ and edges  $E = \{ \text{Pa}_{X^t} \rightarrow  {X^t} \}  $, $\text{Pa}_{X^t}$ denotes the parent nodes of $X^t$ and $\tau_{max}$ denotes the size of the observation window size. $\tau_{max}$ is a hyperparameter in the model training \cite{pamfil2020dynotears}. }
 \end{definition}

 \begin{definition}
 causal temporal adjacency matrix: \textnormal{Causal Temporal Adjacency Matrix (TAM) $\mathcal A$ is the adjacency matrix defined over the causal temporal graph G.}
 \end{definition}

The ideal situation would be $\tau_{max}$ exceeds the maximum lag between time series variables. These definitions enables our proposed graph neural network. Note that the temporal causal graph is a sub-component of the full time graph. Since the full time is translational invariant, it can be generated from the causal temporal graph.

\subsection{Causal Discovery for Multi-Variate Time Series Data}

Consider a $d$-variate time series $(X^t)_{t\in\mathbb{Z}}=(x_1^t,\cdots,x_d^t)$, where the variables influence each other in a time-lagged manner. A general structural causal model (SCM) for the stochastic process can be described as
\begin{align}
    x_i^t=f_i(\text{Pa}_i^{\tau_{max}},\cdots,\text{Pa}_i^1,z_i^t)
\end{align}
where $\text{Pa}_i^{\tau}$ denotes a set of variables that influence $x_i^t$ at $t-\tau$. $z_i^t$ are jointly independent noise terms, $i=1,\cdots,m, t\in\mathbb{Z}, \tau=1,\cdots, \tau_{max}$. $\tau_{max}$ is the maximal lag of the causal temporal graph.

\subsubsection{General SCM for Multi-variate Time Series Data}
A special case is that the model is a class of VAR model with additive noise. Given $m$ variables $\bm{x}\in\mathbb{R}^d$. For a time series of length $T$, we use $a_{ij}^{\tau}$ to denote the impact of $\bm{x}_j$ on $\bm{x}_i$ with a time lag of $\tau$.
\begin{align}
    \bm{x}_i^t=\sum\limits_j\sum\limits_{\tau}a_{ij}^{\tau}\bm{x}_j^{t-\tau}+\bm{z}_i^t
\end{align}
The variable $\bm{x}$ can be a $d$-dimensional vector, or reduced to a scalar when $d=1$. $\bm{z}$ is the random noise of the same shape with $\bm{x}$, typically independent Gaussian exogenous factors. By taking the matrix form,
\begin{align}
X^t=\sum\limits_{\tau} A^{\tau}X^{t-\tau}+Z^t
\end{align}
where $X^t=[\bm{x}_1^t~|\cdots|~\bm{x}_m^t]\in\mathbb{R}^{m\times d}, Z^t\in\mathbb{R}^{m\times d}$ is the noise matrix, $A^{\tau}=(a_{ij}^{\tau})\in\mathbb{R}^{m\times m}$ is introduced to describe the causal relation between variables, and the compact form can be derived as
\begin{align}\label{eq6}
X^t=\mathcal{A}\mathcal{X}+Z^t
\end{align}
where $\mathcal{X}=[X^{t-1\top}~|\cdots|~X^{t-p\top}]^\top\in\mathbb{R}^{pm\times d}, \mathcal{A}=[A^1~|\cdots|~A^p]\in\mathbb{R}^{m\times pm}$ corresponds to the TAM.

For a generalized case we have,
\begin{align}
    X_j^t=f( \text{Pa}(X_j^t) ,Z_j^t)
\end{align}
$f$ is an arbitrary function on parents node $X_j^t$ and independent noise. We only
formulate the evolution of the time series with a specific time slice from its past observation window. The reason
for this simplification comes from the fact that normally the time series is lengthy and sliding window is a reasonable approach for this type of problem.

\subsubsection{GNN-based SCM}

Graph neural network is known for its phenomenal power in representation learning for graph. The nodes in the time series contains abundant information from the past, and hence we might utilize the message passing scheme of GNN and approximate complex functions for the dynamics.  We parametrize the neural network for SCM along with a graph neural networks,
\begin{align}
X^t=f_{\mathcal{A}}(\mathcal{X}, \, Z^t) \label{noise_form}
\end{align}
where $f_{\mathcal{A}}$ is a graph neural network and $\mathcal A$ parametrizes the temporal adjacency matrix which encodes the causal graph, where the non-zero elements suggest the existence of causal relation between the corresponding variables.

\section{Framework of GDBN}
We propose a bayesian deep generative model with a graph neural network to capture the inter connections among time series nodes. The dependency relations can be described with message passing mechanism in a graph neural network \cite{kipf2018neural} \cite{lowe2020amortized}. Instead of treating the graph edges as hidden variables, we allow straightforward gradient descent through elements of the temporal adjacency matrix.
\subsection{Time Series Window Construction}
\begin{figure*}[!h]
\centering
  \includegraphics[width=410pt]{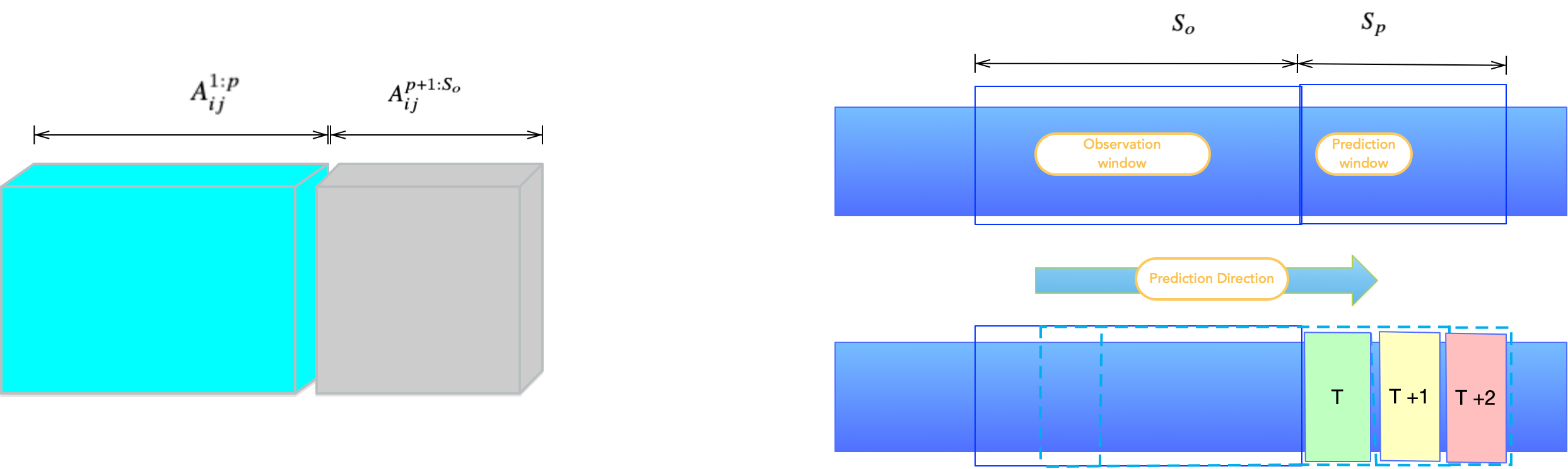}
  \caption{Left: Construction of the Causal Temporal Graph with maximal time length of $S_o$. Each slice of the tensor $A$ represents the connection of $i$th node and $j$th node with a certain lag. Right: Reconstruction of the time window. Given the observation window of length $s_o$, at every step we predict next value of the time series based on recent past $s_o$ values. After $s_p$ steps, the reconstruction is completed for the current time window. The dotted squares are rolling observation windows from left to right which denotes the direction for recurrent prediction.}
  \label{recons}
\end{figure*}

Now we build the sliding window in the time series, which consists of two parts, one for observation followed by the other one for prediction. By setting the observation window size $s_o$ and the prediction window size $s_p$, we assume that the current effect to the future in time series $\bm{x}$ lasts at most $s_o$ steps, and rebuild the following time series of $s_p$ steps is reasonable. If $s_o>p$, we expect all learned $A^{p+1},\cdots,A^{s_o}$ to be zero matrices. Please see Figure \ref{recons} for detail.


\subsection{Model Architecture}

Here we propose a neural network for inference of the causal relation among multi-variate time series data. It has two main components. The first one is a proposed graph neural network with message passing scheme following a strict causal relational path. Nodes and edges form the representation of values of the time series and their time difference, e.g. lags. The second component is variational inference network that produce a variational objective for the optimization of the temporal causal graph and structural equation end to end, namely, our framework, is not only capable of achieving causal discovery in a computationally friendly manner, but also provision a tool for prediction purpose of the time series data with strong explainability.

\subsubsection{Proposed Graph Neural Network}

Graph Neural Network (GNN) is a powerful machinery to introduce relational inductive bias. The neural networks learns the node representation by local message passing mechanism via the graph structure. All variants of GNN \cite{GCN} \cite{DBLP:journals/corr/HamiltonYL17} \cite{velikovi2017graph} \cite{graphormer} have been shown effective on graph structure data related tasks such as graph classification, node classification, link prediction and etc.. Following \cite{gilmer2017neural}, the local message passing mechanism between vertices (v) and edges (e) can be summarized in the following equations: 

\begin{align}
v \rightarrow e: &\quad h^l_{(i, j)} = f^l_e ([h^l_i, h^l_j,x_{(i,j)}])\label{eq:message_passing1}\\
e \rightarrow v: &\quad h^{l+1}_j = f^l_v ([\sum_{i \in \mathcal{N}_j}h^l_{(i, j)}, x_j])\label{eq:message_passing}
\end{align}

$i$ and $j$ denote the node in the graph. $x$ represents the raw feature, while $h^l$ represents the embedding in layer $l$.
For example, $x_{(i,j)}$ is the raw feature of the edge $i \rightarrow j$ and $h^l_j$ is the $l$th embedding layer of the node $j$. $\mathcal{N}_j$ represents the set of neighboring nodes for node $j$. The square bracket denotes the concatenation of the features. The skip connection\cite{he2016identity} with raw feature of $x_{(i,j)}$ in (\ref{eq:message_passing1}) or $x_j$ in (\ref{eq:message_passing}) is optional. $f_e$ and $f_v$ can be modeled with neural networks. \cite{GCN} and \cite{GAT}, including the GNN we had proposed can be treated as special case of (\ref{eq:message_passing}). 

\hfill

{\it Proposed Graph Neural Network} ($\mathscr{G}$)

\begin{itemize}
    \item  Node Embedding (NE): NE maps the node feature to a new embedding space. 

    \item Edge Neural Network (ENN): edge feature can be attained with a edge neural network of a non-linearized transformation. For example, we can choose to encode the lag $\tau$ as edge feature. We aggregate the node embedding and edge embedding by concatenation to update the embedding of edge, followed by an element-wise multiplication operation with the temporal adjacency matrix $\mathcal{A}$. 

    \item Node Neural Network (NNN): NNN maps the aggregated edge features back to the node again. 
\end{itemize}
Notice that the propagation of the messages follows a strict topological order which is faithful to the temporal order. Considering the causal effect from the node $j$ at $t'$ to node $i$ at $t$, $\tau=t-t'=1,\cdots, p$, $\mathscr{G}$ can be formulated as
\begin{align*}
    \text{NE}:  &\quad h_{it}=f_{emb}(x_{it})\\
    \text{ENN}: &\quad h^l_{(it, jt')} = A_{ij}^{\tau}f^l_e ([h^l_{it}, f_\tau(\tau)])\\
    \text{NNN}: &\quad h^{l+1}_{it} = f^l_v (\sum_{j,t'} h^l_{(it, jt')})
\end{align*}
$f_{emb}, f_{\tau}, f_e, f_v$ are parameterized by the neural networks, which take the form of Multi-Layer Perceptron (MLP) in this work. See Algorithm \ref{GNN} for the module of computing GDBN.

\begin{algorithm}
	\renewcommand{\algorithmicrequire}{\textbf{Input:}}
	\renewcommand{\algorithmicensure}{\textbf{Output:}}
	\caption{GNN for Temporal Causal Graph ({\bf GENC}) that updates the node embedding}
	\label{alg}
	\begin{algorithmic}

	    \REQUIRE 
	    Time series of length $X_{1:m}^{i:i+s_o-1}$\\
	    Temporal Adjacency matrix $\mathcal{A}$
	    
	    EMBED: MLP on feature dimension

	    \FOR {each layer $j$ of the Graph Neural Network}
            \STATE Update node embedding $h^j_o \leftarrow$ EMBED$(X_{1:m}^{i:i+s_o-1})$, $\forall i$
            \STATE Update edge embedding $h^j_\tau \leftarrow$ EMBED$(\tau)$, $\tau$ represents the lag between two nodes.
            \STATE Update edge embedding $h^j_e \leftarrow$ EMBED($h^j_o \bigoplus h^j_\tau))$
            \STATE Update $(X_{1:m}^{i:i+s_o-1}) \leftarrow h^j_n \leftarrow$ EMBED$(\mathcal{A}h^j_e)$
        \ENDFOR
        
        \RETURN $\mathcal A$, $X_{1:m}^{'s_o + 1: s_o+s_p}$, $M$, $\log\Sigma$

	\end{algorithmic}  
	\label{GNN}
\end{algorithm}

\hfill

\subsubsection{Variational Autoencoder}

\begin{figure*}[!h]
\centering
  \includegraphics[width=450pt]{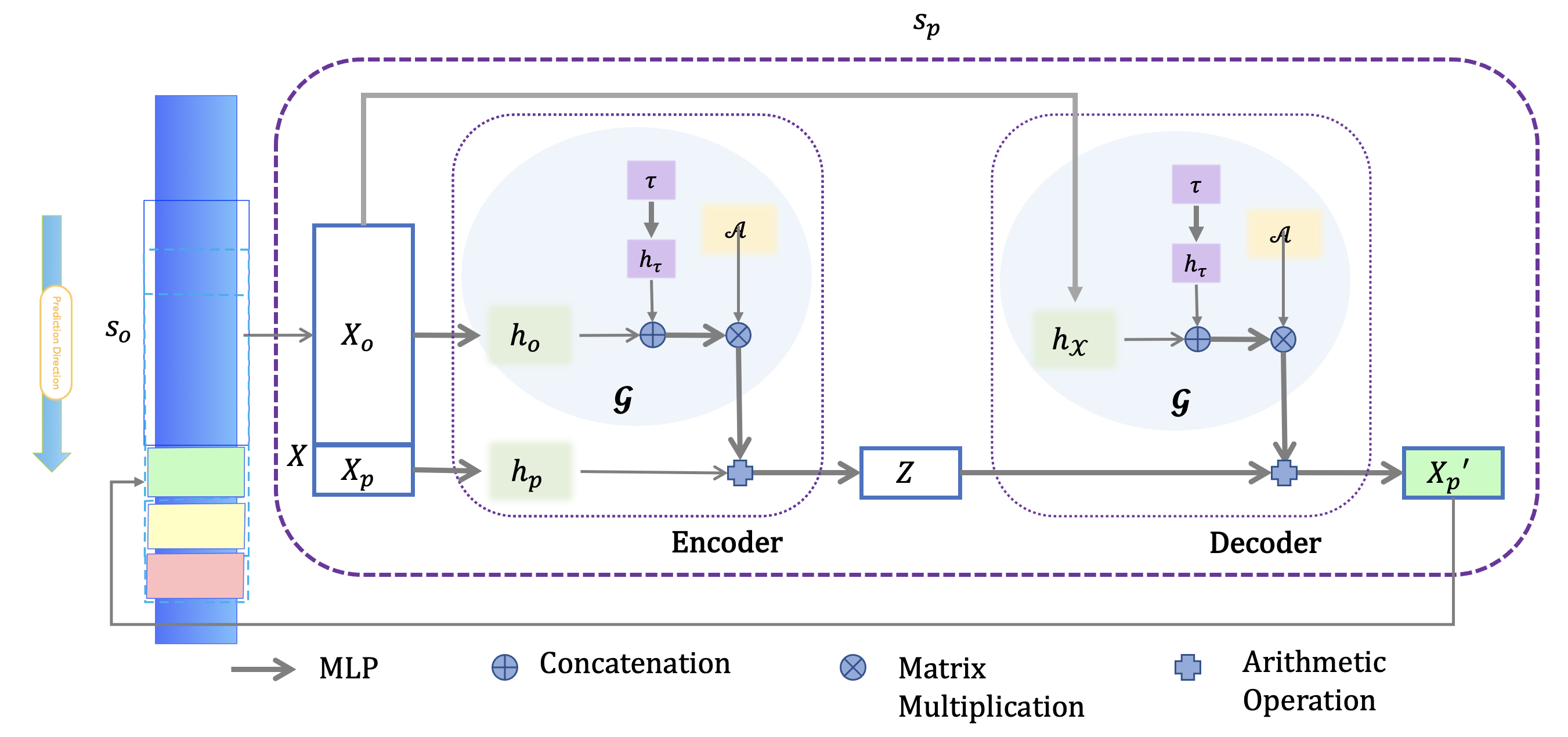} 
  \caption{Model architecture of GDBN. Time series of length $s_o+1$ is fed into the model and the output is the prediction $X_p'$. At time $t$, $X_p, X_o$ are $X^t,\mathcal{X}$ respectively. The rounded rectangle is a repeated unit for $s_p$ times. $h_\tau$ is the embedding of raw edge feature $\tau$.}
\end{figure*}

For each step of reconstruction of the time window as in Figure \ref{recons}, the noise can be generalized from linear Granger model in (\ref{noise_form}) with our proposed GNN $\mathscr{G}$. We assume they yield independent multi-variate Gaussian distribution. At the same time, prediction window is computed with a decoding procedure faithful to the causal graph. The encoder and decoder are
\begin{align}
[ \mu_{Z^t} | \log{\sigma_{Z^t}} ] &=  F_1 \big( F_2(X^t) -  \mathscr{G} (\mathcal{X}) \big) \label{eq9}\\
 [ \mu_{X'^t} | \log{\sigma_{X'^t}} ] &= F_3 ( \mathscr{G} (\mathcal{X}) + F_4(Z^t) ) \label{G_decoder}
\end{align}
respectively, where $Z^t$ is the hidden variable in the VAE model, $F_i, i=1,\cdots,4$ can be either an identity map or an MLP layer that acts upon the feature dimension.


We used a recurrent decoder that can forecast for multi steps based on one time decoder described above while the reconstruction loss and the KL-divergence for multi-steps are jointly trained. During the training, we model $P_{\mathscr{G}}(X_{t+1}| X'_t, X_{t-1} .. X)$, where $X'_t$ as predicted from (\ref{G_decoder}). The observation and prediction window described in Figure \ref{recons} slides forward along with the causal path. We found in the training process, it is a good choice to use only the expectation value $\mu_{X'_t}$ instead of the sampled value for successive recurrent prediction.  This has been proved to be effective for the causal learning as shown in experiments section. For the detail of GDBN network, see a summarized Algorithm \ref{GDBN}

\begin{algorithm}
	\renewcommand{\algorithmicrequire}{\textbf{Input:}}
	\renewcommand{\algorithmicensure}{\textbf{Output:}}
	\caption{Forward GDBN Network Module (\bf GDBN)}
	\label{alg}
	\begin{algorithmic}
	    \REQUIRE 
	    Number of observed variables: $m$\\
	    Size of the observation window: $s_o$\\
	    Size of the prediction window: $s_p$\\
	    Time series of length $s_o+s_p$: $X_{1:m}^{1:s_o+s_p}$\\
	    EMB: MLP on feature dimension
	    \ENSURE  TAM $\mathcal A$, Reconstructed Series $X_{1:m}^{'s_o + 1: s_o+s_p}$, $Z_{1:m}^{s_o + 1: s_o+s_p}$
	   \FOR {$i\leftarrow 1$ to $s_p$}
        \STATE ENCODE:
        \STATE \quad logits $\leftarrow$ EMB(EMB$(X_{1:m}^{i+s_o}$)-GENC($X_{1:m}^{i:i+s_o-1})$)
        \STATE \quad $M, \log\Sigma \leftarrow$ EMB$(logits)$
        \STATE \quad $Z_{1:m}^{i+s_o} \leftarrow$ SAMPLE$(M, \log\Sigma)$
        \STATE DECODE: 
        \STATE \quad $X_{1:m}^{'i+s_o} \leftarrow$ EMB(GENC$(X_{1:m}^{i:i+s_o-1}) +$ EMB $(Z_{1:m}^{i+s_o}))$
        \STATE REPLACE: $X_{1:m}^{i+s_o} \leftarrow X_{1:m}^{'i+s_o}$
        \ENDFOR
        \RETURN $\mathcal A$, $X_{1:m}^{'s_o + 1: s_o+s_p}$, $M$, $\log\Sigma$
	\end{algorithmic}  
	\label{GDBN}
\end{algorithm}

\subsubsection{Optimization Objective}

{\bf Evidence Lower Bound}

Since the exact inference is intractable, the Evidence Lower Bound (ELBO) is optimized in order to approximate the posterior distribution,
\begin{align}
\text{ELBO}(\theta,\phi;X)=&\mathbb{E}_{q_{\phi}(Z|X)}\left[\log p_{\theta}(X|Z)\right] \nonumber \\
&-KL(q_{\phi}(Z|X)||p_{\theta}( Z))\label{LB}
\end{align}

where $p_{\theta}(Z)$ is the prior distribution of the latent variables, $p_{\theta}(X|Z)$ is the conditional distribution of observed $X$ given $Z$ and $\theta$ are the true parameters. $q_\phi$ is the variational posterior parameterized by $\phi$.


For the inference network, the variational posterior is a factored Gaussian with mean $M_Z\in\mathbb{R}^{m\times d}$ and variance $\Sigma_Z\in\mathbb{R}^{m\times d}$.
For the generative network, the likelihood $p(X|Z)$ is a factored Gaussian with mean $M_X\in\mathbb{R}^{m\times d}$ and variance $\Sigma_X\in\mathbb{R}^{m\times d}$.
Then the complexity loss of the latent representation is derived as

\begin{align}
  &-KL(q_{\phi}(Z|X)||p_{\theta}( Z)) \nonumber \nonumber \\
  =&\frac{1}{2}\sum\limits_{i=1}^m\sum\limits_{j=1}^d\left(1+\log(\Sigma_Z)_{ij}-(M_Z)_{ij}^2-(\Sigma_Z)_{ij}\right) \label{KL_loss}
\end{align}
and the reconstruction loss can be computed with Monte Carlo approximation:
\begin{align}
&\mathbb{E}_{q_{\phi}(Z)}\left[\log p_{\theta}(X|Z)\right] \approx \nonumber \\  &-\frac{1}{L}\sum\limits_{l=1}^L\sum\limits_{i=1}^m\sum\limits_{j=1}^d\left(\frac{(X_{ij}-(M_x^l)_{ij})^2}{2(\Sigma_X^l)_{ij}}+\log(\Sigma_X^l)_{ij}+c\right) \label{Recons_Loss}
\end{align}
where $L$ denotes the number of Monte Carlo samples, $c$ is a constant.

\hfill

\noindent {\bf Sparsity and Acyclicity Constraint}  

We add $L1$-norm penalty for the temporal adjacency matrix as a standard approach to introduce sparsity in the optimization objective. Since, we have ignored inter-slice dependencies on the temporal graph, causal relations suffices to have strict topological order and hence no acyclicity constraint needs to be enforced. However, our method can be naturally adapted to inter-slice relations by adding acyclicity constraint in the inter-slice temporal adjacency graph.

\subsection{Training}\label{sec:thre}

Now that we have all the components for optimization, the training process is composed of the following,

\begin{itemize}
    \item Given training example $X=X^{1:s_o+s_p}$, we first run the encoder and compute $q(Z | X)$, then we sample $Z$ and utilize the decoder to compute the prediction values for $X^{s_o+1},\cdots,X^{s_o+s_p}$. 

    \item We compute the ELBO objective with estimated KL loss and reconstruction error in (\ref{KL_loss}) and (\ref{Recons_Loss}). We add the regularized term for sparsity constraint and the optimization objective is estimated,
   \begin{align}\label{eq2}
    \min \limits_{\Phi} \min\limits_{\mathcal{A}\in \mathbb{R}^{m\times pm}}\frac{1}{n}\sum\limits_{n}l_{\mathcal{A},\Phi}(X)+\lambda ||\mathcal{A}||_1
    \end{align}
    where $n$ is the number of samples, $\lambda$ is the coefficient of sparsity and $l$ is the opposite of ELBO defined in (\ref{LB}). $\Phi$ denotes the parameters of the neural network. 

    \item The code is implemented in Pytorch. We perform gradient descents through the neural network parameters and $\mathcal{A}$.  When the optimization process finishes, a hard threshold $\omega$\label{symbol} on $\mathcal{A}$ is set. Any weight smaller than $\omega$ in absolute value is treated a vanishing matrix element, which provably reduces the number of false discoveries\cite{zhou2009thresholding}. 

\end{itemize}

Training process of the network is given in \ref{GDBN-NN},

\begin{algorithm}
	\renewcommand{\algorithmicrequire}{\textbf{Input:}}
	\renewcommand{\algorithmicensure}{\textbf{Output:}}
	\caption{Causal Inference for causal temporal graph}
	\label{alg}
	\begin{algorithmic}
	    \REQUIRE  
        Number of observed variables: $m$\\
	    Size of the observation window: $s_o$\\
	    Size of the prediction window: $s_p$\\
	    
	    \ENSURE Adjacency Matrix $\mathcal A$
	    \REPEAT
	        \STATE Sample time series of length $s_o+s_p$: $X_{1:m}^{1:s_o+s_p}$ from the whole time series $X$
	        \STATE Compute $\mathcal A$, $X_{1:m}^{'s_o + 1: s_o+s_p}$, $M$, $\log\Sigma$ = GDBN($X_{1:m}^{1:s_o+s_p}$, $\mathcal A$)
            \STATE Compute ELBO$(\theta, \phi. \mathcal A)$ with $M$, $\log\Sigma$, $X_{1:m}^{1:s_o+s_p}$ from (\ref{KL_loss}) and (\ref{Recons_Loss}).
            \STATE Compute the sparsity term $||\mathcal A||_1$
            \STATE Compute the gradients $g \leftarrow \nabla$ $ELBO + \lambda ||\mathcal A||_1$
            \STATE $\theta, \phi, \mathcal A$ $\leftarrow$ Update parameters using gradients
        \UNTIL{Convergence of $\theta, \phi, \mathcal A$}
        \RETURN $\mathcal A$
	\end{algorithmic}  
	\label{GDBN-NN}
\end{algorithm}

\section{Experiments}

\subsection{Datasets, Baselines and Performance Metrics}

Since we are evaluating the performance of our inference network for causal discovery. We focus on two datasets that has validation ground truth of causal links.

\subsubsection{Dataset}

\paragraph{Our own synthetic Dataset} We build our own synthetic dataset following the two steps:

\noindent {\it Temporal Adjacency Matrix Generation}

The temporal adjacency matrix $\mathcal{A}=[A^1~|\cdots|~A^p]\in\mathbb{R}^{m\times pm}$ suffices to generate the full time series data given initial condition. We first assign the positions of nonzero elements. In order for the causal temporal graph to meet the assumptions mentioned in section \ref{s1}, $\mathcal{A}$ has to satisfy Hypothesis \ref{hyp:first} and Hypothesis \ref{hyp:second}. Hence among all possible positions for nonzero elements, several are assigned nonzero values in a random way. We set the ratio of zero elements in the temporal adjacency matrix $r\in[0,1]$, to control the sparsity of the causal graph. Besides, about half of the final nonzero positions are assigned with negative values. Then we sample the weights of $\mathcal{A}$ uniformly with absolute values in the range [0.7, 0.95], as suggested by Mastakouri et al.\cite{pmlr-v139-mastakouri21a}.

\hfill

\noindent {\it Linear Dataset and Nonlinear Dataset Generation}

Now we simulate the time series with TAM $\mathcal{A}$ based on (\ref{eq6}). For the linear case, we set a time series of the $m$-variable vector $\bm{x}$, which corresponds to a $p$th order Vector Autoregressive (VAR) model:
\begin{align}\label{eq4}
    \bm{x}^t=A_1\bm{x}^{t-1}+\cdots+A_p\bm{x}^{t-p}+\bm{\epsilon}^t
\end{align}
where $A$ is an $m$th-order parameters matrix, $\bm \epsilon^t$ is an independent noise, $\bm \epsilon\sim \mathcal{N}(\bm 0, \sigma I)$. To guarantee the stability, all roots of $\det(I-A_1z-\cdots-A_pz^p)$ are limited to lying outside the unit circle. See (\ref{stationarity}) for details. Notice that the variables are scalar-valued for simplicity but can be easily generalized to the vector-valued ones.

For the nonlinear case, we consider two ways:

\begin{align}
X^t=f(\mathcal{A}\mathcal{X})+Z \label{eq7}\\
X^t=\mathcal{A}g(\mathcal{X})+Z \label{eq8}
\end{align}
where the nonlinear function $f,g$ perform elementwise. (\ref{eq7}) is a widely adopted form considering nonlinearity. We also notice that (\ref{eq8}) will not change the causal graph structurally, as Yu et al.\cite{yu2019dag} reason its property with Taylor expansion. Here we consider $f(\cdot)=g(\cdot)=\sin(\cdot)$ as the nonlinear form for convenience.

With $X^1, \cdots, X^p$ initialized independently in a random way, the time series of length $T$ can be computed by generating $X^t$ at each time step $t$ repeatedly for $t=p,\cdots,T$. 
We construct time series samples of length $s$ with sliding windows.

\paragraph{Other Benchmark Dataset} To further test our proposed model, we also adopt the dataset generated by \cite{Lawrence2021DataGP}, which provides a system to generate time series data with flexible configurations for causal discovery. Notice the data is different from ours in that it allows for a varying likelihood of nonlinear functions given a TAM. We set the maximum lag $p=2$, the number of variables $d=5$ and generate TAM under out assumptions, then simulate time series with gaussian noise. Here we adopt the concept of complexity, which is set with different default levels in \cite{Lawrence2021DataGP}. For level 0, the time series is generated only in a linear way. Increasing level stands for increasing percentage and variety of non-linear dependencies, including piecewise linear, monotonic and periodic functions.

\subsubsection{Baselines}

To validate the proposed model GDBN, we compare its performance with other popular approaches for bayesian networks inference. Our baseline is PCMCI \cite{runge2019detecting}, PCMCI+\cite{runge2020discovering} and DYNOTEARS \cite{pamfil2020dynotears} without intra-slice acyclicity constraint according to our assumptions about the time series data. PCMCI is strong baseline based on conditional independence test. It is compatible with our assumption of Causal stationarity, no contemporaneous causal links and no hidden variables. DYNOTEARS is similar to ours that the optimization objective is the same, i.e., the adjacency matrix of DAG.

\subsubsection{Performance Metrics}
The inference results are evaluated on some common graph metrics: 1) False discovery rate (FDR), 2) True positive rate (TPR). 3) F1 computed as the harmonic mean of (1-FDR) and TPR, 4) Structural Hamming Distance (SHD), which is the number of required changes to the graph to match the ground truth and computed as the sum of missing edges and extra edges. In view of the time order, the direction of all edges are already determined according to our assumptions. True (T) is the set of true edges, False (F) is the set of non-edges in the ground truth graph. Positive (P) is the set of estimated edges, where True Positive (TP) is an edge that is in the true graph, while False Positive (FP) is not.
\begin{itemize}
    \item FDR = \# FP / \# P
    \item TPR = \# TP / \# T
    \item F1 = 2 $\times$ (1 - FDR) $\times$ TPR / (1 - FDR + TPR)
    \item SHD = \# P + \# T - 2 $\times$ \# TP
\end{itemize}

Considering these metrics relies on the threshold set manually(\ref{sec:thre}), we also introduce AUROC.

\subsection{Results and Hyperparameters}

We set the maximum lag $p=5$, the number of variables $d=10$ to simulate the nonlinear time series of length $T=600$. We set the coefficient for L-1 norm $\lambda=0.01$ for GDBN and DYNOTEARS, regularization parameter $\alpha_{\text{PC}}=0.1$ for PCMCI, and threshold are tuned for each configuration to achieve a desirable F1 score. For GDBN, there are more hyperparameters. we set the observation window size $s_o=10$, the prediction window size $s_p=3$, the feature dimension of the latent variables $d_Z=8$, and the number of hidden units for MLP $h=32$. The performances are shown in Table \ref{tab_o} for the nonlinear data (\ref{eq7}), and Table \ref{tab_i} for the nonlinear data (\ref{eq8}). Note that we run 5 repeated random experiments and take an average to report the performances of GDBN and DYNOTEARS, while for PCMCI, which is significantly time-consuming, we only run once.



\begin{table*}[!h]
		\caption{Performances of the methods for the nonlinear data (\ref{eq7})}\label{tab_o}
		\centering
		\begin{tabular}{ccccccc}
			\toprule
			\#Variables & Methods & FDR & TPR & F1 & SHD & AUC \\
			\midrule
			\multirow{3}*{10} & DYNOTEARS & 0.056 & 0.880 & 0.910 & 9.40 & 0.973$\pm$ 0.026\\
			~ & PCMCI & 0.302 & 0.545 & 0.612 & 38.0 & 0.845\\
			~ & PCMCI+ & 0.030 & 0.582 & 0.727 & 24.0 & 0.863\\
			~ & GDBN & 0.020 & 0.964 & 0.972 & 3.00 & \textbf{0.996$\pm$ 0.009}\\
			\midrule
			\multirow{3}*{15} & DYNOTEARS & 0.035 & 0.813 & 0.882 & 26.0 & 0.986$\pm$ 0.009\\
			~ & PCMCI & 0.080	& 0.675	& 0.779	& 46.0 & 0.938 \\
			~ & PCMCI+ & 0.012 & 0.683 & 0.808 & 39.0 & 0.942\\
			~ & GDBN & 0.000    & 0.975   & 0.987  & 3.00 & \textbf{0.998$\pm$ 0.004}\\
			\midrule
			\multirow{3}*{20} & DYNOTEARS & 0.079 & 0.537 & 0.678 & 107  & 0.930$\pm$ 0.014\\
			~ & PCMCI & 0.010 & 0.476 & 0.643 & 111 & 0.913\\
			~ & PCMCI+ & 0.061 & 0.510 & 0.660 & 110 & 0.910\\
			~ & GDBN & 0.000    & 0.779   & 0.874  & 46.4  & \textbf{0.983$\pm$ 0.018}\\
			\bottomrule
		\end{tabular}
	\end{table*}
	
	\begin{table*}[!h]
		\caption{Performances of the methods for the nonlinear data (\ref{eq8})}\label{tab_i}
		\centering
		\begin{tabular}{ccccccc}
			\toprule
			\#Variables & Methods & FDR & TPR & F1 & SHD & AUC \\
			\midrule
			\multirow{3}*{10} & DYNOTEARS & 0.588  & 0.371 & 0.390 & 63.6 & 0.838$\pm$ 0.026\\
			~ & PCMCI & 0.535 &	0.727 & 0.567	& 61.0 & 0.859\\
			~ & PCMCI+ & 0.370 & 0.527 & 0.574	& 43.0 & 0.900\\
			~ & GDBN & 0.000  & 1.000  & 1.000  & 0.00  & \textbf{1.000 $\pm$ 0.000 }\\
			\midrule
			\multirow{3}*{15} & DYNOTEARS & 0.774 & 0.270 & 0.246 & 199 & 0.732$\pm$ 0.009\\
			~ & PCMCI & 0.543	& 0.708	& 0.556	& 136 & 0.900 \\
			~ & PCMCI+ & 0.071 & 0.658 & 0.771	& 47.0 & 0.923\\
			~ & GDBN & 0.000  & 1.000  & 1.000  & 0.00  & \textbf{1.000 $\pm$ 0.000 }\\
			\midrule
			\multirow{3}*{20} & DYNOTEARS & 0.830 & 0.175 & 0.173 & 352  & 0.677$\pm$ 0.014\\
			~ & PCMCI & 0.508	& 0.557	& 0.522	& 214 & 0.830\\
			~ & PCMCI+ & 0.140 & 0.529 & 0.655	& 117 & 0.871\\
			~ & GDBN & 0.000 & 1.000  & 1.000  & 0.00  & \textbf{1.000 $\pm$ 0.000 }\\
			\bottomrule
		\end{tabular}
	\end{table*}

GDBN outperforms other methods, especially in the detection of complicated nonlinear causal relations. And GDBN has a particular advantage when applied to relatively large datasets with high accuracy and efficiency. 

\begin{figure}[!h]  
    \centering
		\subfigure[]{
		    \label{fig_1}
			\centering
			\includegraphics[width=0.2\textwidth]{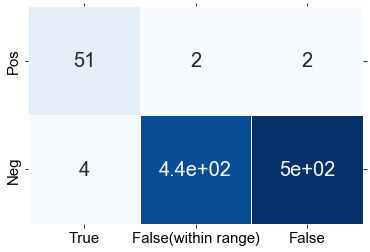}
		}
		\subfigure[]{
		    \label{fig_2}
			\centering
			\includegraphics[width=0.2\textwidth]{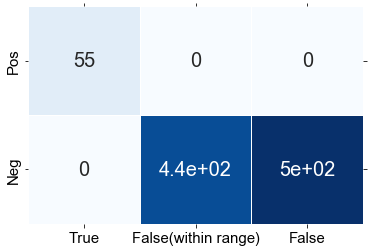}
		}
		\subfigure[]{
		    \label{fig_3}
			\centering
			\includegraphics[width=0.2\textwidth]{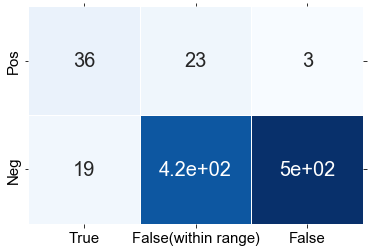}
		}
		\subfigure[]{
		    \label{fig_4}
			\centering
			\includegraphics[width=0.2\textwidth]{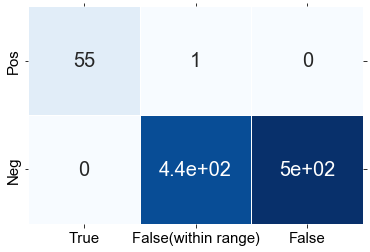}
		}
	\caption{Comparison: confusion matrix of DYNOTEARS(Figure(a), (c)) and GDBN(Figure(b), (d)). In the estimated causal temporal graph $G_e$, edges identified by a model are denoted as \emph{Pos}, or \emph{Neg} otherwise. Since the observation window size $o_w$ is larger than the true maximum lag $p$, some edges in $G_e$ do not exist in the ground truth graph $G_t$, which are denoted as \emph{False}. In $G_t$, \emph{True} denotes existing edges, \emph{False(with range)} otherwise. For (a), (b), the time series are generated in the nonlinear way (\ref{eq7}) while (c), (d) in another way (\ref{eq8}).} \label{cm} 
\end{figure}

\begin{figure*}[!h]  
    \centering
		\subfigure[]{
			\centering
			\includegraphics[width=0.2\textwidth]{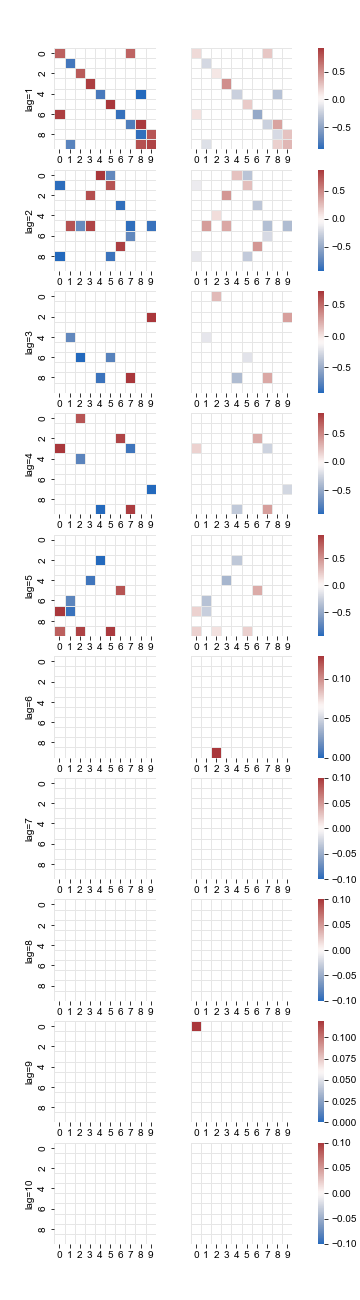}
		}
		\subfigure[]{
			\centering
			\includegraphics[width=0.2\textwidth]{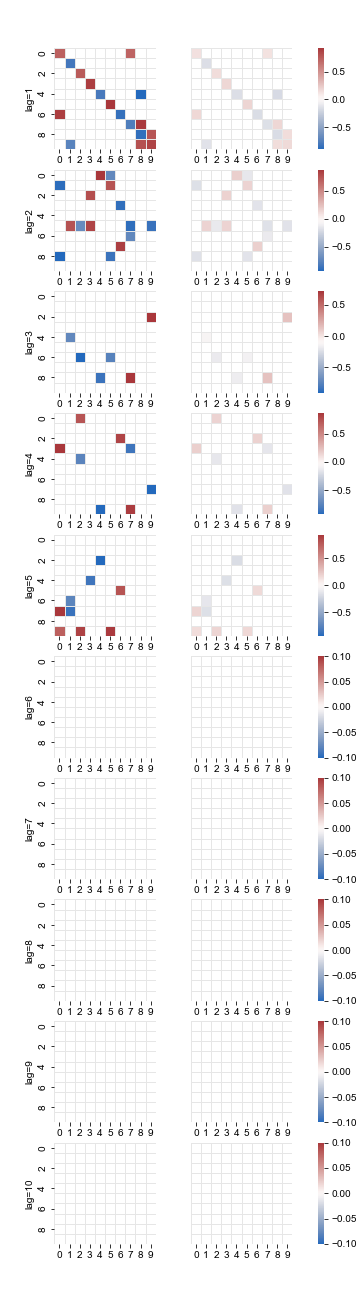}
		}
		\subfigure[]{
			\centering
			\includegraphics[width=0.2\textwidth]{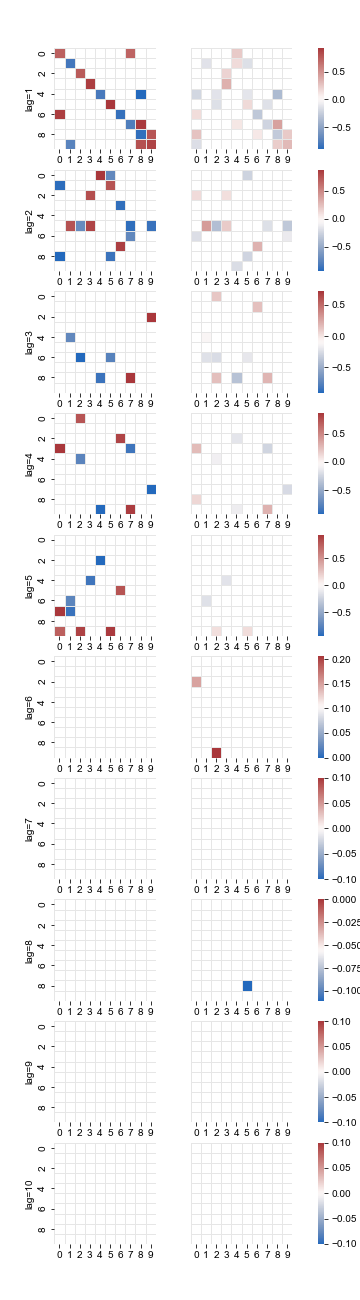}
		}
		\subfigure[]{
			\centering
			\includegraphics[width=0.2\textwidth]{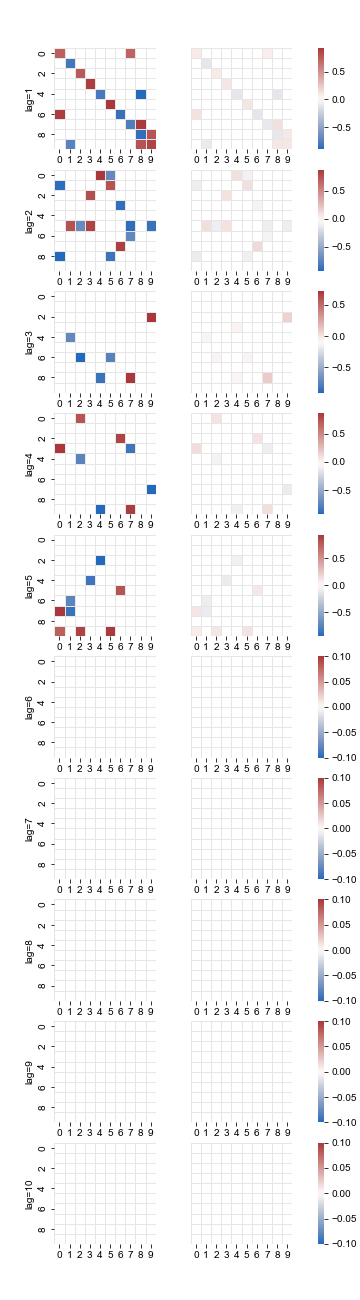}
		}
	\caption{Comparison: the causal temporal graph of DYNOTEARS (Figure(a), (c)) and GDBN (Figure(b), (d)). In each subfigure, the ground truth causal temporal graph is showed on the left while the estimated truth one on the right. For (a), (b), the time series are generated in the nonlinear way (\ref{eq7}) while (c), (d) in another way (\ref{eq8}).} \label{vis_m}
\end{figure*}
For nonlinear data, GDBN gives an outstanding performance, especially decreasing the number of false discoveries within range of the maximum lag $p$ when compared with DYNOTEARS, shown in Figure \ref{cm} and Figure \ref{vis_m}. This shows the effectiveness of GDBN on nonlinear data and DYNOTEARS is generally not sufficient for nonlinear scenario. 

\begin{table}[H]
\centering
\caption{AUC with increasing complexity level}\label{tab:exp_dg}
\begin{tabular}{ccccc} 
\toprule 
Complexity & DYNOTEARS & PCMCI & PCMCI+ & GDBN   \\ 
\midrule
0  & 0.955   & 1.000  & 0.86 & \textbf{1.000}  \\
10 & 0.864   & 0.871  & 0.833 & \textbf{0.943}  \\
20 & 0.909   & \textbf{0.951}  & 0.86 & 0.946  \\
30 & 0.727   & \textbf{0.915}  & 0.86 & 0.856  \\
\bottomrule
\end{tabular}
\end{table}
We show the GDBN performance on dataset in \cite{Lawrence2021DataGP} for different level of complexity in Table \ref{tab:exp_dg}. In case of large number of edges, GDBN reaches comparable perforance with saving a signifcant ammount of time shown in Table \ref{tab:timecost}. The experiment is carried on  Intel(R) Core(TM) i7-8557U CPU @ 1.70GHz. The time recorded for GDBN is for 100 epochs so that GDBN reaches stable AUC performance. 

\begin{table}[H]
\centering
\caption{Average Time Cost of the methods for Experiments in Table \ref{tab:exp_dg}. All experiments are run on the same device. For GDBN, we count the convergence time. }\label{tab:timecost}
\begin{tabular}{ccccc} 
\toprule
Method & DYNOTEARS & PCMCI & PCMCI+ & GDBN   \\ 
\midrule
Time (s) & 0.009 & 1829 & 204.8 & 53.48\\
\bottomrule
\end{tabular}
\end{table}
\section{Conclusion and Discussion}


In this work, we introduce GDBN, a method that can infer causal dependencies for a full time graph in a discrete multi-variate temporal graph. As a  preliminary, we define a fundamental structure of temporal graph and the relevant adjacency graph matrix. We formalize the inference network to causal discovery and structural causal model, and designed the way to train the network jointly from end to end. We conducted experiments and benchmark our methods on two main different types of time series data with the known ground truth. Our approach reaches significantly higher precision than popular benchmarks such as Dynotears\cite{pamfil2020dynotears} and PCMCI \cite{runge2019detecting}. Our method can be utilized widely in the real world problems, as understanding causal relational knowledge is becoming a pressing need for decision makers.

In despite of the success of our method, the  graph neural networks of higher expressive power is expected. The question remains regarding what type of graph neural network suites the causal discovery is suitable for general SCM defined in \cite{zheng2020learning}. The other important problem is what if our time series data is downsampled or irregularly sampled from the sufficient time series data and how our framework can be adjusted to adapt to such cases.

\bibliographystyle{plain}
\bibliography{sample-base}

\end{document}